\documentclass[sigconf]{acmart}
\usepackage{algorithm}
\usepackage{algorithmicx}
\usepackage{algpseudocode}
\usepackage{multirow}
\usepackage{tabularx} % 引入tabularx包
\AtBeginDocument{%
  }

\setcopyright{acmlicensed}
\copyrightyear{2018}
\acmYear{2018}
\acmDOI{XXXX.XXXX}
\acmConference[Conference KDD 'XX]{}{June 03--05,
  2018}{Woodstock, NY}
\begin{document}

\title{DP-DGAD: A Generalist Dynamic Graph Anomaly Detector with Dynamic Prototypes}

% \author{Anonymous authors}
% \affiliation{%
% \renewcommand{\shortauthors}{Anonymous authors}
% \renewcommand{\shorttitle}{DP-DGAD: A Generalist Dynamic Graph Anomaly Detector}
% \affiliation{%

% \affiliation{%
\author{Jialun Zheng}
\affiliation{%
  \institution{Department of Computing}
  \institution{The Hong Kong Polytechnic University}
  \city{Hong Kong}
  \country{China}
}
%\email{22069255r@connect.polyu.hk}
\author{Jie Liu}
\affiliation{%
  \institution{Department of Computer Science}
  \institution{City University of Hong Kong}
  \city{Hong Kong}
  \country{China}
}
%\email{jliu288@cityu.edu.hk}
\author{Jiannong Cao}
\affiliation{%
  \institution{Department of Computing}
  \institution{The Hong Kong Polytechnic University}
  \city{Hong Kong}
  \country{China}
}
%\email{csjcao@comp.polyu.edu.hk}
\author{Xiao Wang}
\affiliation{%
  \institution{School of Software}
  \institution{Beihang University}
  \city{Beijing}
  \country{China}
}
%\email{xiao_wang@buaa.edu.cn}
% Add emails after all authors
\author{Hanchen Yang}
\authornote{Corresponding author.}
\affiliation{%
  \institution{Department of Computing}
  \institution{The Hong Kong Polytechnic University}
  \city{Hong Kong}
  \country{China}
}
%\email{hanchen.yang@connect.polyu.hk}

\author{Yankai Chen}
\affiliation{%
  \institution{Department of Computer Science}
  \institution{University of Illinons at Chicago}
  \city{Chicago}
  \country{United States}
}
%\email{}

\author{Philip S. Yu}
\affiliation{%
  \institution{Department of Computer Science}
  \institution{University of Illinons at Chicago}
  \city{Chicago}
  \country{United States}
}
%\email{psyu@uic.edu}

\renewcommand{\shortauthors}{Zheng et al.}

\begin{abstract}
Dynamic graph anomaly detection (DGAD) is essential for identifying anomalies in evolving graphs across domains such as finance, traffic, and social networks. Recently, generalist graph anomaly detection (GAD) models have shown promising results. They are pretrained on multiple source datasets and generalize across domains. While effective on static graphs, they struggle to capture evolving anomalies in dynamic graphs. Moreover, the continuous emergence of new domains and the lack of labeled data further challenge generalist DGAD. Effective cross-domain DGAD requires both domain-specific and domain-agnostic anomalous patterns. Importantly, these patterns evolve temporally within and across domains. Building on these insights, we propose a DGAD model with Dynamic Prototypes (DP) to capture evolving domain-specific and domain-agnostic patterns.
Firstly, DP-DGAD extracts dynamic prototypes, i.e., evolving representations of normal and anomalous patterns, from temporal ego-graphs and stores them in a memory buffer. The buffer is selectively updated to retain general, domain-agnostic patterns while incorporating new domain-specific ones. Then, an anomaly scorer compares incoming data with dynamic prototypes to flag both general and domain-specific anomalies. Finally, DP-DGAD employs confidence-based pseudo-labeling for effective self-supervised adaptation in target domains. Extensive experiments demonstrate state-of-the-art performance across ten real-world datasets from different domains. 
\end{abstract}

%    <concept>
%        <concept_id>10010147.10010257</concept_id>
%        <concept_desc>Computing methodologies~Machine learning</concept_desc>
%        <concept_significance>500</concept_significance>
%        </concept>
%  </ccs2012>




\maketitle

\section{Introduction}
Graph anomaly detection (GAD) identifies unusual nodes or edges based on their feature distributions. It has wide applications in fraud detection, financial transaction analysis, and social networks~\cite{deng2022graph,wang2025epm, pan2025label}. However, real-world graphs are often evolving~\cite{zheng2024inductive}, such as new accounts being added or existing profiles being updated in the social networks. Unlike static GAD, dynamic graph anomaly detection (DGAD) focuses on capturing not only anomalous patterns but also their temporal evolution. Therefore DGAD is a more challenging yet more representative of real-world scenarios compared with the setting of GAD~\cite{liu2024multivariate, kong2024causalformer,10.1145/3748259}.

\begin{figure}[]
\centering
\includegraphics[width=8.5cm]{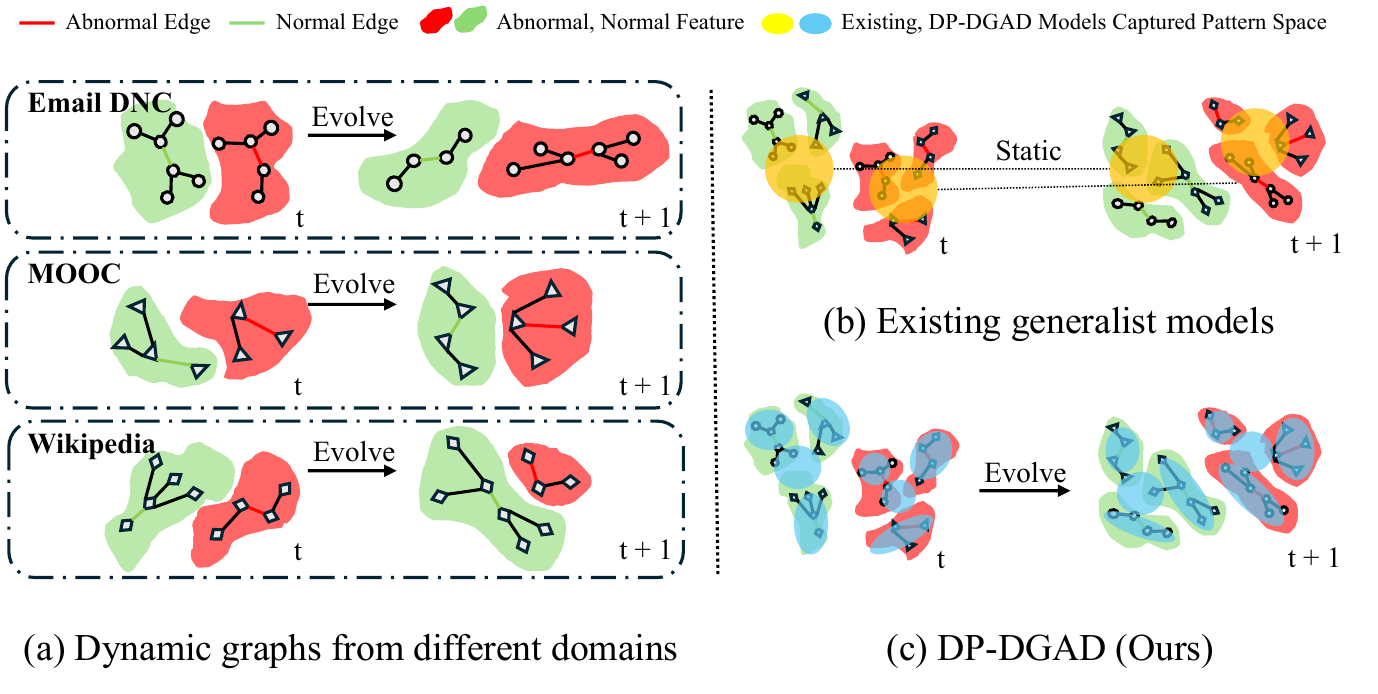}
\caption{(a) Anomalous patterns change over time. (b) Existing generalist methods are static, capturing only domain-agnostic patterns. (c) Our method, DP-DGAD, captures both domain-agnostic and domain-specific patterns while adapting dynamically over time. } 
\label{teaser}
\end{figure}

With the advancement of deep learning, several methods have been developed to address DGAD problems~\cite{wang2025epm}. Early work, such as TADDY~\cite{liu2021anomaly}, employs transformers to capture both structural and temporal patterns for anomaly detection. More recently, approaches like GeneralDyG~\cite{yang2025generalizable} and FALCON~\cite{chen2024fine} have introduced unsupervised DGAD models. They assume all training datasets have only normal samples, and detect anomalies by identifying samples that deviate from normal distributions. 
% On top of that, SLADE~\cite{lee2024slade} make even more strict self-supervised learning setting of zero labeling information include normal labels and generate pseudo labels to guide model update.
Despite promising results, existing DGAD models follow a “one model per domain” paradigm, limiting generalization to new domains. However, datasets from different domains have distinct anomalous patterns, feature distributions, and structures, leading to domain shift~\cite{qiaogcal,tian2024freedyg}. It hinders the generalization capability of current DGAD models across datasets from different domains. Consequently, there is a growing demand for a unified DGAD model that can adapt to multiple datasets from different domains.

Recently, generalist models for static graphs, such as ARC~\cite{liu2024arc} and AnomalyGFM~\cite{qiao2025anomalygfm}, have emerged as promising solutions. These generalist models are pretrained on source datasets and generalizing to target datasets from different domains. However, these models cannot be directly applied to DGAD due to the following challenges: \textbf{1. Temporally evolving anomalous pattern.} Current generalist detectors for static graphs assume static spatial anomalies, neglecting the temporal dynamics of evolving graphs. In contrast, anomalous patterns in dynamic graphs constantly change, such as node and edge feature distributions and structures (see Fig.~\ref{teaser}(a)). These changes significantly complicates generalization.  \textbf{2. Continuous domain shift}. The continuous emergence of new domains presents a challenge: models must balance domain-specific and general patterns, often sacrificing one for the other (as shown in Fig.~\ref{teaser}(b). This results in unstable performance across datasets. \textbf{3. Lack of labeled datasets.} Existing methods struggle to incorporate new patterns from target datasets due to the lack of labeled data, which is labor costly. As a result, they rely only on patterns from source datasets, which may differ from those in target datasets and cause performance drops.

To tackle above challenges, we introduce DP-DGAD, a generalist Dynamic Graph Anomaly Detection method based on Dynamic Prototypes. Here dynamic prototypes refer to representations reflect the abnormal and normal classes, which evolve over time. The basic idea is illustrated in Fig.~\ref{teaser}(c). DP-GDAD captures both domain-specific and domain-agnostic patterns via dynamic prototypes as they evolve over time. These patterns futher help to mitigate negative effects from continuously emerging domains 

Specifically, our approach generates dynamic prototypes by first capturing spatial anomalies from ego-graphs. These dynamic prototypes are updated with edge-level temporal attributes, creating representations that capture the evolution of anomalous patterns.
Afterward, dynamic prototypes are saved in a memory buffer, which strategically retains shared, domain-agnostic prototypes and integrates new, domain-specific ones when encountering new domains. This mechanism enables DP-DGAD to adapt to new domains while retaining generalizability, thereby effectively countering continuous domain shifts. Furthermore, the learned prototypes themselves function as effective anomaly scoring for detecting anomalies. This intrinsic property ensures confident detection with low entropy loss for pseudo-labeling. Using pseudo labels, DP-DGAD can adapt to unlabeled target domains~\cite{tian2024self}. This ability to perform adaptation without groundtruth anomaly labels makes DP-DGAD a practical and powerful solution for real-world scenarios that lack of labels. Extensive experiments on ten real-world datasets validate the state-of-the-art performance of DP-DGAD.
The key contributions are summarized as follows:

\begin{itemize}
\item We, for the first time, formally define and address the generalist DGAD problem. The problem aims to train a DGAD model on multiple labeled source datasets from different domains and generalize to unlabeled target datasets.  
% \item We identify three critical challenges to solve the aforementioned problem, namely, the need to generalize both spatial and temporal anomalous patterns in dynamic graphs, continuous domain shift among datasets, and the inability to learn new anomalous patterns in the target domain due to a lack of labels. 
\item We introduce DP-DGAD, a novel generalist DGAD model based on Dynamic Prototypes. Our approach extracts and updates prototypes that capture evolving spatial and temporal patterns. We selectively retain prototypes relevant to new domains to ensure balanced and adaptive learning. Then, it leverages high-confidence detections from target datasets as pseudo-labels, enabling fully self-supervised adaptation to target domains. 
\item Experiments show our method achieves the best performance among all advanced baselines on ten cross-domain datasets. Our code is now available\footnote{https://anonymous.4open.science/r/KDD2026-DP-DGAD-30C4}.

\end{itemize}

\section{Related Work}
\subsection{Dynamic Graph Anomaly Detection}

Traditional Graph Anomaly Detection (GAD) \cite{ma2024graph} aims to identify nodes or edges with abnormal patterns. Notwithstanding, it struggles with evolving graphs, where node features, edge attributes, and structures change over time. Dynamic Graph Anomaly Detection (DGAD)~\cite{liu2024multivariate} addresses this limitation and is applied in inherently dynamic fields such as traffic~\cite{deng2022graph} and social networks~\cite{wang2025epm}. To capture these evolving patterns, several deep learning methods have been proposed. For example, TADDY \cite{liu2021anomaly} uses transformers to encode spatial and temporal changes. Followed by StrGNN and DynAnom \cite{cai2021structural, guo2022subset} that focus on local subgraphs to efficiently detect structural anomalies. Additionally, STGAN \cite{deng2022graph} introduces a convolutional adversarial network to handle evolving anomalies, including structural and feature changes. Alternatively, models like SAD and CoLA \cite{tian2023sad, liu2021anomaly} use semi-supervised learning with partial labels. However, aforementioned approaches cannot be directly applied real world scenarios due to lack of labeled datasets.

Unsupervised DGAD \cite{yang2025generalizable}, which operates without labels, has gained attention for its practicality. Most unsupervised DGAD methods assume training data contains only normal samples and learn node or edge embeddings to model this normal distribution. Among them, FALCON \cite{chen2024fine} leverages fine-grained temporal information through enhanced sampling and embedding. Then it improve generalization via an attention alignment optimized by contrastive learning loss \cite{zhu2024topology, wu2025retrieval}. Instead of focusing on embedding, SLADE \cite{lee2024slade} models evolving long-term interactions to learn normal patterns and find anomalies. However, these methods typically follow a "one model per domain" paradigm and struggle to generalize across different domains.
\subsection{Generalist Anomaly Detector}

First explored in image anomaly detection, the generalist anomaly detector paradigm involves pretraining models on multiple source datasets to generalize across different domains \cite{zhu2024toward, xu2025towards}. To generalize learned patterns to new domains, the model must capture invariant anomalous patterns \cite{carvalho2023invariant, wang2023generalist, chen2025filter} across domains. However, compared to image data, graph data exhibit more complex and variable structures that also change across domains \cite{ han2023intra,yang2023higrn,yang2024uniocean}.

Recently, generalist approaches for GAD \cite{liu2024arc, qiao2025anomalygfm, niu2024zero} have emerged as a promising direction due to their cross-domain generalizability. These methods commonly use unified strategies, such as normalization \cite{zhao2024all, li2023zero}, prototype-based approaches \cite{wang2024unveiling, sun2024program, wan2024federated} or prompt-based \cite{liu2023graphprompt} method to address domain differences. Although results are promising, their unified strategies often lead to unbalanced performance, excelling on some datasets while performing poorly on others. Moreover, they are not suited for dynamic graphs, which require capturing anomalous patterns in an evolving manner. 
\begin{figure*}[t] 
	
	\centering
	
	\includegraphics[width=18cm]{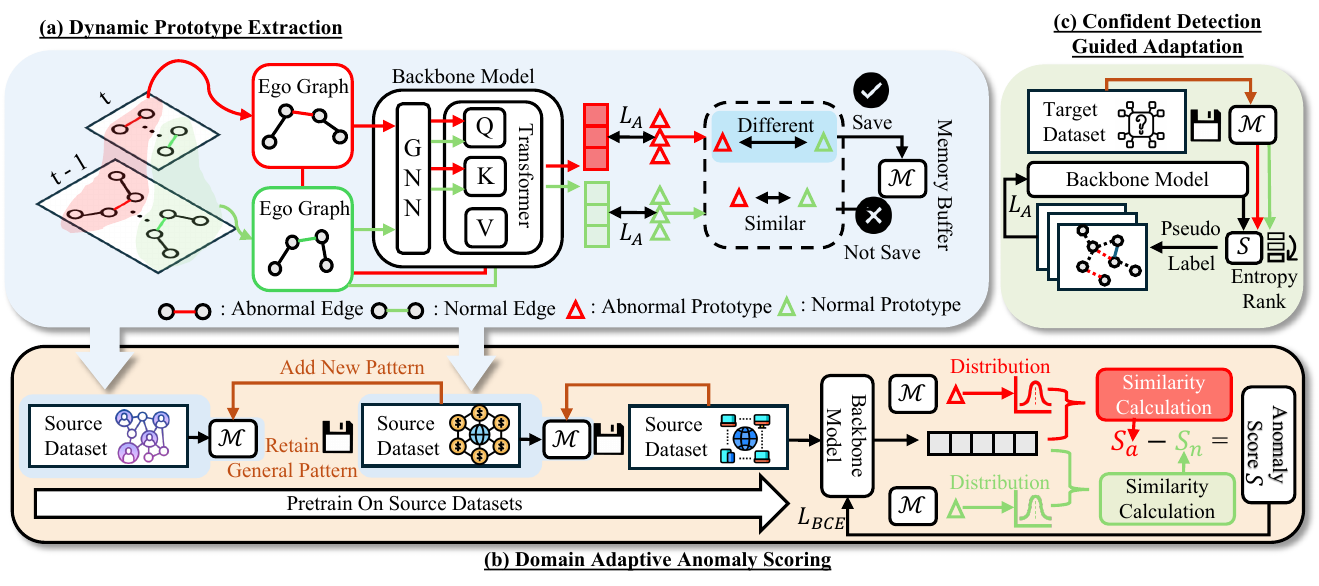}

	     \caption{\textbf{General framework of DP-DGAD}. (a) Starting with the first source dataset, we extract ego-graphs to capture temporal patterns and store the most distinct prototype pairs in a memory buffer. (b) As we pretrain on the following source datasets, new domain-specific patterns are added while general domain agnostic patterns are retained; prototype distributions are then compared with edge embeddings for anomaly scoring. (c) On target datasets, the pretrained anomaly scorer selects confident low-entropy detections for pseudo-labeling and model updating, aligning prototypes with the target domain.}

	\label{frame}
	
\end{figure*}

\section{Preliminaries}

\textbf{Notations.} Dynamic graph can be denoted as $\mathcal{G} = \{G_1, G_2, .... G_T\}$ with $T$ being number of time intervals, and each interval consists $M$ number of timestamps. Specifically, for each time interval $t \in \{1,2,...,T\}$, we have $G_t \in \{G_1, G_2, .... G_T\}$. We further have $G_t = (V_t, E_t)$, $V_t$ is a node set consisting of entities, $E_t$ is an edge set consisting of spatial relations. Additionally, $A^{V_t} \in \mathbb{R}^{N^{V_t} \times N^{V_t}}$ and $A^{E_t} \in \mathbb{R}^{N^{E_t} \times N^{E_t}}$ are the node adjacency matrix and edge adjacency matrix during time interval $t$, respectively, where $N^{V_t} = \vert V_t\vert$, $N^{E_t} = \vert E_t\vert$ are the number of nodes and edges. We further have normal edge set $E^n_t$ and abnormal edge set $E^a_t$, where each edges have label $y\in \{0,1\}^{N^{E_t}}$ with $0$ denoting the normal edges and $1$ denoting the abnormal edges. During certain time interval $t$, nodes and edges will record data as features that can be represented as $X^{V_t} \in \mathbb{R}^{N^{V_t} \times D^V \times M}$, $X^{E_t} \in \mathbb{R}^{N^{E_t} \times D^E \times M}$ where $D^V$ and $D^E$ are the dimension of nodes and edges' data, respectively.

\noindent\textbf{Continuous Domain Shift.} Given $\{\mathcal{G}_1, \mathcal{G}_2, \ldots, \mathcal{G}_{S+K}\}$ being a collection of $(S+K)$ dynamic graph from different domains, where $\{\mathcal{G}_{1}, \mathcal{G}_{2}, \ldots, \mathcal{G}_{S}\}$ are from source domain and $\{\mathcal{G}_{S+1}, \mathcal{G}_{S+2}, \ldots, \mathcal{G}_{S+K}\}$ are from target domains. Each domain's dynamic graph $\mathcal{G}_{i}$ can be represented as $\mathcal{G}_{i} = \{G_1, G_2, \ldots, G_{T_i}\}$ with $T_i$ time intervals. For two dynamic graphs from different domains, $\mathcal{G}_{i}$ and $\mathcal{G}_{j}$ where $i \neq j$, we define domain shift as the difference in distributions of node and edge features, graph structural such as the adjacency matrix or the mapping from edges to anomaly labels. This domain shift continuously occurs between any two dynamic graphs in the dataset collection $\{\mathcal{G}_1, \mathcal{G}_2, \ldots, \mathcal{G}_{S+K}\}$.

\noindent \textbf{Problem Definition.} This work aimed to develop a generalist anomaly detector to detect anomalous edges at each timestamp within a specified time interval. On top of it, the detector is trained on labeled source datasets, and then generalizes to unlabeled target datasets.
Specifically, given $S$ number of source datasets, our goal is to learn an anomaly scoring function $\Psi$ such that for a specific target dynamic graph $G_k = (V_k, E_k)$, we can predict the label set $y_k$ of $E_k$ using function $\Psi$:

\begin{equation}
\Psi(\mathcal{G}_1, \mathcal{G}_2, \ldots, \mathcal{G}_{S}) = y_k.
\end{equation}

 Notably,  the above function $\Psi$ is trained using labels exclusively from the source datasets, while the target datasets remain unlabeled to simulate a real-world scenario.

\section{Methodology}

DP-DGAD aims to capture both general, domain-agnostic, and domain-specific evolving anomalous patterns. This allows patterns learned from labeled source datasets to generalize to unlabeled target datasets across different domains. Specifically, as shown in Fig.~\ref{frame}, DP-DGAD first extracts evolving patterns using dynamic prototype extraction. Then, during domain-adaptive anomaly scoring, it retains general patterns shared among pretrained source datasets while incorporating domain-specific ones. These dynamic prototypes are compared with embeddings to calculate anomaly scores. The pretrained anomaly scoring module then detects confident anomalous and normal edges with low entropy on the target datasets. These confident detections provide pseudo-labels for unlabeled edges and enable model updating on the target datasets through prototype alignment loss.
\subsection{Dynamic Prototype Extraction}
In this section, our goal is to extract dynamic prototypes as the evolving representations of normal and anomalous patterns. Particularly, we align prototypes with both abnormal and normal patterns to fully exploiting anomaly discriminability. This approach enables a more effective capture of evolving anomalous patterns over time.

For each edge $e_i$, its anomalous pattern can be captured by ego-graph consisting of its neighbors, thereby reflecting how it deviates from or resembles them~\cite{qiao2025anomalygfm}. To capture not only the anomalous pattern but also its evolution over time, temporal ego-graphs are utilized. These graphs~\cite{yang2025generalizable} extract $k$-hop neighboring edges occurring on or before timestamp $t$ around edge $e_i$ and form a subgraph $G_{ego}$. The temporal ego-graph is then input into a simple backbone model consisting of a Graph Neural Network (GNN), followed by a transformer to retrieve representations. The GNN first outputs representation $H_l \in \mathbb{R}^{|E_{ego}|\times d}$ as follows:

\begin{equation}
H_{l} = \sigma  (A_{ego}H_{l-1}W_1^l + H_{l-1}W_2^l),
\end{equation}\label{eq1}

\noindent where $d$ is the dimension of representation and $l$ is the GNN layer index. $W_2^l, W_1^l \in \mathbb{R}^{ D^{l-1}\times D^{l}}$ are learnable parameter matrix of layer $l$ and $\sigma$ is the activation function. $A_{ego}$ refer to the edge adjacency matrix of ego-graph. To make sure the representation captures domain agnostic features, a residual module is adapted here to process edge-specific representation $h_i \in H_l$ as follows:

\begin{equation}
h_i-\frac{1}{|E_{ego}|} \sum_{e_j\in E_{ego}}h_j.
\end{equation}\label{eq2}

Then, the processed $H_l$ is fed into a transformer. This transformer's role is to identify the similarity between edge $e_i$ and its k-hop neighbors. In this way, we encode the spatial anomalous patterns of $e_i$ as they evolve over time into its embedding $z_i$.

\begin{equation}
z_i = \sum_{h_j\in H_l}\frac{exp(\frac{<w_Qh_i, w_Kh_j>}{\sqrt{d_{out}}})}{\sum_{h_j\in H_l}exp(\frac{<w_Qh_i, w_Kh_j>}{\sqrt{d_{out}}})}w_Vh_j,
\end{equation}\label{eq3}

\noindent where $<.,.>$ denotes the dot product, $d_{out}$ refers to the dimension of the output representation $z_i$ of edge $e_i$. $w_Q, w_K, w_V$ refer to the Query, Key, and Value matrix. 

After obtaining the representation of the each edges, we align dynamic prototypes with both abnormal and normal edges' representation. Formally, we have $p_n \in \mathbb{R}^{d_p}$ as the dynamic normal prototypes while $p_a\in \mathbb{R}^{d_p}$ represents the dynamic abnormal prototypes. Both for abnormal $e_i$ in $E^a_t$ and that in $E^n_t$, we can obtain the ego-graph representation. We denote the representation of abnormal edges as $Z_a$ and the representations of normal edges as $Z_n$. To align $p_a$ and $p_n$ with $Z_a$ and $Z_n$, we define alignment loss function:

\begin{equation}
L_A = \sum^{|E_t|}_{i=1}I_{y_i=0}||z_i-p_n||^2_2 + I_{y_i=1}||z_i-p_a||^2_2,
\end{equation}\label{eq5}

\noindent where $I_{y_i=0}$ represents an indicator function. This function returns a value of 1 when the condition $y_i=0$ is met, and 0 otherwise.

These aligned dynamic prototypes are further stored in dynamic prototype buffer $\mathcal{B}$ that has size $\mathcal{M}$, set as 10\% of training dataset size. The buffer is updated during each iteration to capture diverse aspects of the abnormal and normal patterns as possible. Due to limitations in size and memory, we retains only the most discriminative prototypes. This is achieved by ranking all prototypes currently in the buffer in ascending order based on difference score. The prototypes that are found to be the least distinct (i.e., most similar to others) are then replaced. The difference score is calculated as the mean Euclidean distance between pairs of prototypes:

\begin{equation}
s_d= \frac{1}{{d_p}} \sum_{m=1}^{{d_p}} ||p_{a}^m - p_{n}^m||_2,
\end{equation}\label{eq6}

\noindent where $d_p$ refer to the dimension of prototypes, $p_{a}^m, p_{n}^m$ refer to the corresponding dimension $m$'s feature in prototypes. In each training epoch, the system identifies the pair of prototypes that exhibit the highest difference score. This most distinct pair is then used to set the initial values for $p_n$ and $p_a$ for that specific epoch. This strategy ensures that the dynamic prototypes are continuously refined to be more distinguishable.

\subsection{Domain Adaptive Anomaly Scoring}

Distribution-based anomaly scoring \cite{guo2024infproto} surpasses traditional binary classifiers in its ability to generalize across different domains. This is because it captures higher-order statistics of data distributions by comparing data with both abnormal and normal distributions. However, our current dynamic prototypes are domain-specific, which prevents their direct application across different domains. To generalize them to new domains, we identify and save domain-agnostic ones, while adding new prototypes from new domains. Subsequently, we can build an anomaly scoring module that leverages these enhanced prototypes.

For identifying domain-agnostic patterns, they should already be present in the memory buffer, which store information from previously encountered source datasets. Additionally, they should also exhibit similarity to the feature representations found in the subsequent source dataset. The similarity $s_e$ is calculated as the mean Euclidean distance between dynamic prototype pair and the representations of edges in the new domain dataset:

\begin{equation}
s_e=  \frac{1}{|E_t|} \sum_{i=1}^{|E_t|} ||z_i - p_{n}||_2 + ||z_i - p_{a}||_2.
\end{equation}\label{eq7}

Notably, the representation $z_i$ is projected to the same dimension as that of dynamic prototypes.
A lower distance indicates higher similarity, meaning more general prototypes. This will be combined with the difference score in the following manner:

\begin{equation}
s_r=\lambda_d s_d-\lambda_es_e,
\end{equation}\label{eq8}

\noindent where $\lambda_d$ and $\lambda_e$ are parameters used to control the contribution of the two scores. A higher $s_r$ indicates greater pairwise difference and similarity to the new source dataset, representing more general domain-agnostic patterns. For new domain patterns, we rank buffer prototypes by $s_r$ in ascending order and replace the lowest-scoring pair with one from the new domain. We then initialize $p_n$ and $p_a$ using the highest $s_r$ pairs to ensure dynamic prototypes capture both domain-agnostic and domain-specific patterns.

With these dynamic prototypes in the memory buffer, we build an anomaly scoring module by measuring edge representation similarity to both normal and abnormal distributions. Top prototype pairs alone cannot fully represent distributions, so we iteratively update statistical measures (mean and covariance). Each iteration incorporates new data and existing statistics to capture evolving distributions. The mean update as follows:

\begin{equation}
    \mu_{n,t} = \alpha \mu_{n,t-1} + (1-\alpha)\sum^{\mathcal{M}}_{i=1} p_{n,i},
\end{equation}\label{eq9}
\begin{equation}
    \mu_{a,t} = \alpha \mu_{a,t-1} + (1-\alpha)\sum^{\mathcal{M}}_{i=1} p_{a,i},
\end{equation}\label{eq10}

\noindent where $\mu_n$ and $ \mu_a$ are the mean values for normal and abnormal distributions. $\alpha$ is a momentum parameter to control the ratio of updating based on prototypes and existing mean value. 

For covariance updates, we first calculate center embedding, which captures how individual prototypes differ from the overall average. Then, we use center embeddings to update covariance that inherently measures how data points vary together around their mean values:

\begin{equation}
C_n = [p_{n,1} - \mu_{n,t}, p_{n,2} - \mu_{n,t},...,p_{n,\mathcal{M}} - \mu_{n,t}]^T,
\end{equation}\label{eq11}
\begin{equation}
C_a = [p_{a,1} - \mu_{a,t}, p_{a,2} - \mu_{a,t},...,p_{a,\mathcal{M}} - \mu_{a,t}]^T,
\end{equation}\label{eq12}
\begin{equation}
    \Sigma_{n,t} = \Sigma_{n,t-1} + (1-\alpha)\frac{C_n^TC_n}{max(1,\mathcal{M}-1)},
\end{equation}\label{eq13}
\begin{equation}
    \Sigma_{a,t} = \alpha \Sigma_{a,t-1} + (1-\alpha)\frac{C_a^TC_a}{max(1,\mathcal{M}-1)},
\end{equation}\label{eq14}

\noindent where $C_n $ and $ C_a$ are the center embeddings, and, $\Sigma_{n} $ and $ \Sigma_{a}$ are the covariance values.

Then we will calculate similarity score $s_a, s_n$ to abnormal distribution and normal distribution respectively:

\begin{equation}
    s_{n,i} = z_i^T\mu_n - \lambda_n z_i^T\Sigma_nz_i,
\end{equation}\label{eq15}
\begin{equation}
    s_{a,i} = z_i^T\mu_a - \lambda_a z_i^T\Sigma_az_i,
\end{equation}\label{eq16}

\noindent where $\lambda_a,\lambda_n$ is the learnable parameter. Next, we compute the anomaly score $s_i = s_{a,i}-s_{n,i}$, which measures the similarity gap to determine if an edge is closer to the abnormal or normal distribution. This score is then converted into a probability ($p_i$) using a sigmoid function, representing the likelihood of the edge being anomalous. The overall learning objective combines binary cross-entropy loss with the previous alignment loss ($L_A$):

\begin{equation}
    L_{BCE} = -\frac{1}{|E_t|}\sum^{|E_t|}_{i=1}y_i\log (p_i) + (1-y_i)\log (1-p_i),
\end{equation}\label{eq17}
\begin{equation}
    L = \lambda_{BCE}L_{BCE} + \lambda_A L_A,
\end{equation}\label{eq18}

\noindent where $\lambda_A$ and $\lambda_{BCE}$ are parameters controlling the contribution ratio of these two losses to the final loss.
\subsection{Confident Detection Guided Adaptation}

Since target datasets often lack labels, we utilize the pretrained anomaly scoring module to generate pseudo labels. This module can generalizes anomalous patterns to target datasets and help DP-DGAD in better adapting to target domains.

We first input the target data into the pretrained anomaly scoring to obtain the probability $p_i$. However, not all detections are highly confident or close to the ground truth. Therefore, we use the entropy value~\cite{tian2024self} to reflect the model's detection reliability, which is calculated as follows:

\begin{equation}
    \mathcal{H}(p_i) = -p_i\log (p_i+\epsilon) + (1-p_i)\log (1-p_i+\epsilon),
\end{equation}\label{eq19}

\noindent where $\epsilon$ is a small constant added for numerical stability. Lower entropy indicates higher confidence in the model’s classification of sample as normal or abnormal. The top $N_{con}$ abnormal and normal detection pairs with the lowest entropy are selected as confident detections. After that, we use them as pseudo-labels for their corresponding edges.

To inject the domain knowledge of target datasets that captured by confident detections, into our dynamic prototypes, we perform an alignment loss $L_A$ guided update. Since true labels are unavailable for the target datasets, we cannot use the BCE loss for this step; instead, we rely solely on the alignment loss $L_A$, utilizing the generated pseudo-labels from the target datasets for updating model parameters. Then, we use the updated model for final anomaly detection on target test datasets.

\section{Experiments}
\subsection{Experimental Setup}

\noindent \textbf{Datasets.} We use 10 real-world datasets~\cite{liu2021anomaly, lee2024slade, yang2012defining} from various domains for evaluation, and the detailed statistics of datasets are presented in Table~\ref{sta}. Details of datasets are listed in Appendix~\ref{A1}.

\begin{table}[h]
    \caption{Statistics of datasets}
  \centering
  \scalebox{0.86}{
  \small
  \begin{tabular}{ c |c c c c}
\hline\hline
    \textbf{Dataset} & \textbf{Domain} & \textbf{\#Nodes} & \textbf{\#Edges} & \textbf{Avg. Degree} \\
\hline
Wikipedia & Online Collaboration & 9,227 &157,474&34.13 \\
Bitcoin-OTC & Transaction Network & 5,881 &35,588&12.10 \\
Bitcoin-Alpha & Transaction Network & 3,777  &24,173&12.80 \\
Email-DNC & Online Communication & 1,866&39,264&42.08\\
UCI Messages & Online Communication & 1,899&13,838&14.57\\
AS-Topology & Autonomous Systems &34,761&171,420&9.86\\
MOOC & Online Courses & 7,074&411,749 & 116.41\\
Synthetic-Hijack &Email Spam &986 &333,200 &38.11\\
TAX51 &Transaction Network&132,524&467,279&7.05\\
DBLP&Citation Network&25387& 119899&9.44\\
\hline\hline
    \end{tabular}}

\label{sta}
\end{table}

\begin{table*}[ht]
\caption{Overall model performance across eight datasets with three anomaly percentages.}
\label{result_formatted_by_metric}
\renewcommand{\arraystretch}{1.1} % Reduced from 1.2
\centering
\setlength{\tabcolsep}{2pt} % Reduce column spacing (default is usually 6pt)
\resizebox{\textwidth}{!}{%
\begin{tabular}{@{}ccl|ccc|ccc|ccc|ccc|ccc|ccc|ccc|ccc@{}} % Added @{} to remove outer margins
\hline\hline
\multirow{2}{*}{\textbf{Metrics}} & \multicolumn{2}{c|}{\multirow{2}{*}{\textbf{Methods}}} & \multicolumn{3}{c|}{AS-Topology} & \multicolumn{3}{c|}{Bitcoin-Alpha} & \multicolumn{3}{c|}{Email-DNC} & \multicolumn{3}{c|}{UCI Messages} & \multicolumn{3}{c|}{Bitcoin-OTC} & \multicolumn{3}{c|}{TAX51} & \multicolumn{3}{c|}{DBLP} & \multicolumn{3}{c}{Synthetic-Hijack} \\
\cline{4-27}
& & & 1\% & 5\% & 10\% & 1\% & 5\% & 10\% & 1\% & 5\% & 10\% & 1\% & 5\% & 10\% & 1\% & 5\% & 10\% & 1\% & 5\% & 10\% & 1\% & 5\% & 10\% & 1\% & 5\% & 10\% \\
\hline
\multirow{10}{*}{\rotatebox{90}{AUROC}} & \multirow{5}{*}{\rotatebox{90}{DGADs}}
&SAD & 40.45 & 42.27 & 46.43 & 59.03 & 64.12 & 63.77 & 52.66 & 54.02 & 58.82 & 59.63 & 58.80 & 62.44 & 72.21 & 74.59 & 77.13 & 44.50 & 46.12 & 52.26 & 45.70 & 44.24 & 40.97 & 62.51 & 64.09 & 69.52 \\
& &SLADE & 41.09 & 45.56 & 48.28 & 66.78 & 65.05 & 64.93 & 49.34 & 51.67 & 55.51 & 55.72 & 57.50 & 59.82 & 70.17 & 69.20 & 74.77 & 47.59 & 48.64 & 50.96 & 45.01 & 43.22 & 41.67 & 66.67 & 61.70 & 63.02 \\
& &FALCON & 48.05 & 53.43 & 55.16 & 58.77 & 64.80 & 67.55 & 80.33 & 86.21 & 85.74 & 66.95 & 67.65 & 71.78 & 66.17 & 70.86 & 73.07 & 51.23 & 54.57 & 53.20 & 48.56 & 53.77 & 51.21 & 59.19 & 63.99 & 68.48 \\
& &GeneralDyG & 43.45 & 47.73 & 49.93 & 66.60 & 69.83 & 70.23 & 52.76 & 54.88 & 57.21 & 51.09 & 54.85 & 59.74 & 73.95 & 72.14 & 75.54 & 53.15 & 50.41 & 50.39 & 51.73 & 47.48 & 48.87 & 58.15 & 51.09 & 52.94 \\
& &TADDY & 41.32 & 44.68 & 42.16 & 42.27 & 46.53 & 44.42 & 51.20 & 53.74 & 57.18 & 50.25 & 53.97 & 56.85 & 47.82 & 49.02 & 49.77 & 41.37 & 43.66 & 47.39 & 40.97 & 41.55 & 39.69 & 60.76 & 61.20 & 63.27 \\
\cline{2-27}
& \multirow{4}{*}{\rotatebox{90}{Generalists}} &GraphPrompt & 49.60 & 51.21 & 53.47 & 46.12 & 46.75 & 45.33 & 69.30 & 71.04 & 74.45 & 60.74 & 59.28 & 59.26 & 50.66 & 52.79 & 54.33 & 53.48 & 55.72 & 58.61 & 60.64 & 61.07 & 62.33 & 72.16 & 76.28 & 74.19 \\
& &UNPrompt & 52.19 & 53.62 & 57.03 & 45.87 & 47.92 & 48.66 & 70.22 & 74.08 & 73.53 & 58.60 & 59.14 & 60.04 & 47.54 & 52.13 & 53.09 & 54.63 & 55.57 & 57.78 & 62.51 & 64.82 & 64.53 & 72.30 & 75.66 & 76.20 \\
& &AnomalyGFM & 53.97 & 54.76 & 58.23 & 47.84 & 47.72 & 50.21 & 72.90 & 73.44 & 75.62 & 60.76 & 61.82 & 60.87 & 54.30 & 52.72 & 52.45 & 55.94 & 56.46 & 56.24 & 61.51 & 60.66 & 60.49 & 74.34 & 79.37 & 77.31 \\
& &ARC & 54.33 & 55.86 & 60.28 & 41.14 & 43.64 & 45.19 & 71.32 & 77.83 & 73.27 & 58.79 & 60.04 & 59.43 & 48.51 & 53.85 & 50.69 & 48.49 & 54.07 & 58.71 & 62.98 & 67.57 & 63.15 & 71.74 & 74.89 & 75.77 \\
\cline{2-27}
& ~ & \textbf{DP-DGAD} & \textbf{71.27} & \textbf{73.89} & \textbf{75.26} & \textbf{76.56} & \textbf{80.23} & \textbf{81.09} & \textbf{87.46} & \textbf{90.03} & \textbf{88.87} & \textbf{70.19} & \textbf{72.24} & \textbf{75.71} & \textbf{85.54} & \textbf{91.04} & \textbf{90.28} & \textbf{57.24} & \textbf{65.19} & \textbf{63.21} & \textbf{68.66} & \textbf{68.32} & \textbf{67.50} & \textbf{90.09} & \textbf{90.25} & \textbf{91.02} \\
\hline
\multirow{10}{*}{\rotatebox{90}{AUPRC}} & \multirow{5}{*}{\rotatebox{90}{DGADs}}
&  SAD & 6.16 & 13.24 & 11.77 & 6.99 & 9.01 & 10.99 & 13.32 & 19.33 & 29.69 & 12.22 & 20.67 & 25.25 & 7.32 & 12.97 & 18.85 & 5.27 & 9.18 & 10.06 & 3.64 & 5.54 & 9.80 & 12.45 & 17.01 & 26.79 \\
& &  SLADE & 6.10 & 12.27 & 10.54 & 5.05 & 9.78 & 11.41 & 14.76 & 21.22 & 27.95 & 12.77 & 19.87 & 24.02 & 5.11 & 9.32 & 11.04 & 4.95 & 6.71 & 12.29 & 2.12 & 6.90 & 10.14 & 10.10 & 15.21 & 20.67 \\
& &  FALCON & 7.66 & 15.29 & 12.11 & 5.12 & 7.90 & 14.97 & 16.24 & 23.56 & 33.10 & 12.33 & 24.59 & 26.15 & 4.52 & 11.72 & 13.16 & 11.59 & 13.48 & 14.11 & 5.37 & 11.16 & 9.68 & 12.72 & 15.77 & 23.32 \\
& &  GeneralDyG & 6.33 & 20.27 & 22.90 & 5.07 & 5.03 & 10.23 & 12.67 & 15.52 & 30.16 & 14.31 & 21.43 & 22.14 & 8.22 & 14.69 & 19.12 & 5.02 & 4.64 & 9.38 & 3.11 & 7.41 & 8.95 & 13.28 & 18.42 & 28.64 \\
& &  TADDY & 5.99 & 11.50 & 13.22 & 4.41 & 6.95 & 11.13 & 13.06 & 17.33 & 28.90 & 11.55 & 18.49 & 23.27 & 5.14 & 10.78 & 12.65 & 5.41 & 8.95 & 10.13 & 2.89 & 6.42 & 8.17 & 11.09 & 10.54 & 23.48 \\
\cline{2-27}
& \multirow{4}{*}{\rotatebox{90}{Generalists}} & GraphPrompt & 11.56 & 13.05 & 20.96 & 8.48 & 10.79 & 12.07 & 20.83 & 26.97 & 37.74 & 20.05 & 24.57 & 27.43 & 4.32 & 10.02 & 16.89 & 10.53 & 12.35 & 14.86 & 6.27 & 8.57 & 9.15 & 10.97 & 14.43 & 27.62 \\
& &  UNPrompt & 12.06 & 14.63 & 21.21 & 9.54 & 10.70 & 12.52 & 21.06 & 25.18 & 38.01 & 21.40 & 26.01 & 30.35 & 5.15 & 12.70 & 18.49 & 11.21 & 12.66 & 14.39 & 7.92 & 8.55 & 10.90 & 13.28 & 16.08 & 28.73 \\
& &  AnomalyGFM & 13.3 & 15.01 & 22.46 & 7.94 & 13.15 & 12.26 & 22.84 & 28.38 & 40.72 & 21.05 & 25.59 & 31.51 & 4.32 & 12.60 & 19.50 & 11.20 & 14.32 & 13.99 & 6.27 & 10.98 & 11.71 & 11.04 & 11.93 & 31.27 \\
& &  ARC & 12.04 & 19.98 & 26.65 & 12.07 & 11.06 & 13.35 & 16.48 & 28.35 & 41.71 & 19.47 & 24.87 & 29.71 & 6.89 & 14.99 & 20.45 & 8.98 & 8.66 & 15.42 & 8.45 & 9.22 & 10.44 & 15.53 & 16.79 & 25.02 \\
\cline{2-27}
& ~ & \textbf{DP-DGAD} & \textbf{19.14} & \textbf{23.34} & \textbf{28.67} & \textbf{17.40} & \textbf{17.69} & \textbf{18.07} & \textbf{34.18} & \textbf{41.59} & \textbf{45.17} & \textbf{37.56} & \textbf{43.22} & \textbf{45.03} & \textbf{17.45} & \textbf{21.85} & \textbf{24.52} & \textbf{13.06} & \textbf{17.13} & \textbf{22.23} & \textbf{11.17} & \textbf{13.02} & \textbf{13.56} & \textbf{20.17} & \textbf{27.24} & \textbf{35.11} \\
\hline\hline
\end{tabular}%
}
\end{table*}
\noindent \textbf{Evaluation Metrics and Baselines.} We will utilize AUROC and AUPRC, two widely used metrics that have also been adapted in other DGAD models. For both metrics, a higher value
denotes a better performance. 

As for the baselines, we compare DP-DGAD with several recent DGAD models, GeneralDyG~\cite{yang2025generalizable}, FALCON~\cite{chen2024fine}, SLADE~\cite{lee2024slade}, SAD~\cite{tian2023sad}, TADDY~\cite{liu2021anomaly}. Generalists for static graphs such as ARC~\cite{liu2024arc}, AnomalyGFM~\cite{qiao2025anomalygfm}, UNPrompt~\cite{niu2024zero}, and GraphPrompt~\cite{liu2023graphprompt} are also adapted for comparison. All generalist GAD models are adapted to process dynamic graphs by incorporating the same GNN and transformer backbone as DP-DGAD. This modification was made to enable a fair comparison, with all other modules unchanged. Details of baselines are listed in Appendix~\ref{A2}.

\noindent \textbf{Implementation Details.} All methods undergo pretraining on two source datasets, Wikipedia and MOOC~\cite{liu2021anomaly}, chosen because they have available ground truth anomalies. Subsequently, they are fine-tuned on eight target datasets, which contain only normal edges and no abnormal ones. Following this, testing is performed on the same eight target datasets, but with injected anomalies.

Parameters for ego-graph extraction, GNN, and the transformer are aligned with those of GeneralDyG~\cite{yang2025generalizable}. The memory buffer size $\mathcal{M}$ is configured as 10\% of the data size. The number of confident detections for pseudo-labeling, $N_{con}$, is 10\% of the data size. The momentum $\alpha$ is set to 0.9. The value of $p$ is varied at 1\%, 5\%, and 10\%. The loss ratios $\lambda_A$ and $\lambda_{BCE}$ are set to 0.1 and 0.9, respectively, while $\lambda_d$ and $\lambda_{e}$ are set to 0.3 and 0.7.

\subsection{Experimental results}

\begin{figure}[htbp]
\centering
\includegraphics[width=8.5cm]{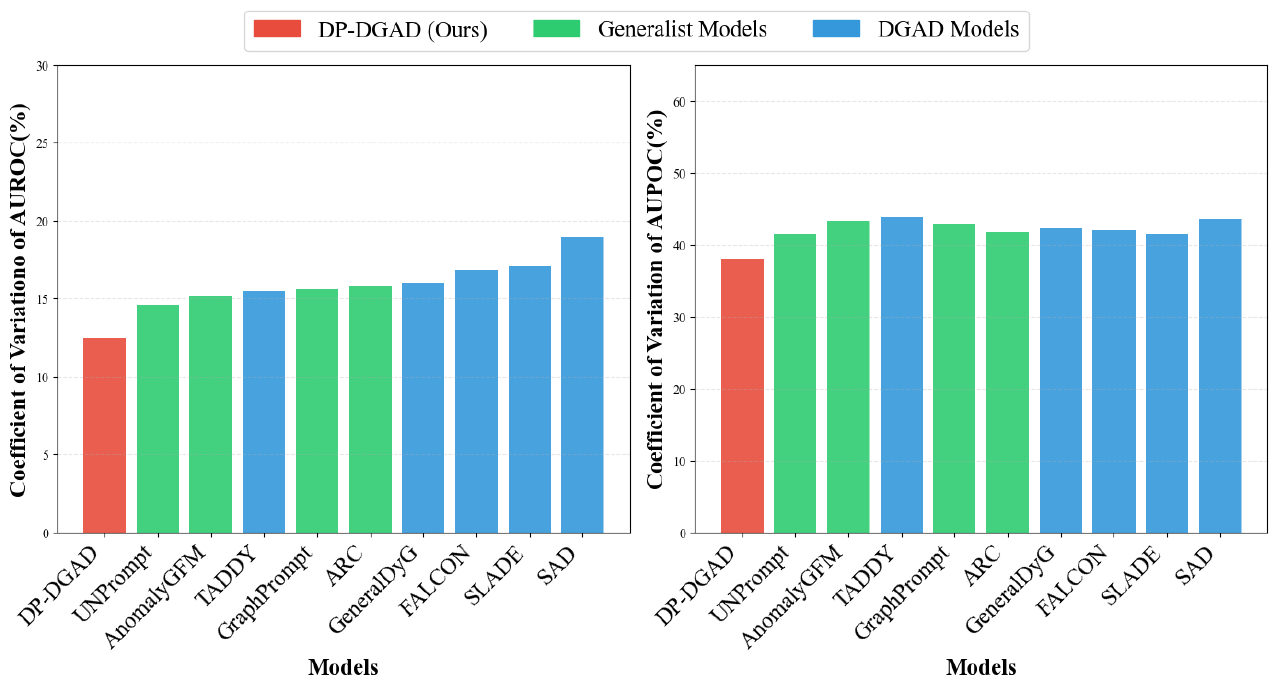}
\caption{Model performance under continuous domain shift. } \label{dev}
\vspace{-0.2cm}
\end{figure}

We compared our method, DP-DGAD, with existing Dynamic Graph Anomaly Detection (DGAD) and generalist Graph Anomaly Detection (GAD) methods across eight target datasets from diverse domains. The comparison results, presented in Table~\ref{result_formatted_by_metric}, lead to several key observations: \textbf{(1) DP-DGAD consistently achieves the highest AUROC and AUPRC among all baselines across the eight datasets.}
For instance, on Email-DNC, DP-DGAD reached an AUPRC of 34.18, significantly outperforming the second best, AnomalyGFM, at 22.84. 
These improvements are substantial, particularly in complex datasets such as TAX51 and DBLP and under low anomaly ratios scenarios.
DP-DGAD's state-of-the-art performance, especially at low anomaly ratios, is attributed to its memory buffer retaining domain-agnostic samples. This buffer enables our model to review anomalous patterns, thus mitigating insufficient anomaly data. \textbf{(2) DP-DGAD achieves stable performance under continuous domain shifts across datasets.} As shown in Fig.~\ref{dev}, DP-DGAD exhibits the lowest coefficient of variation, indicating superior stability. Generalist models, while adapting to capture domain-agnostic patterns, don't match DP-DGAD's performance due to neglecting domain-specific and evolving patterns. However, they are generally better than DGAD models that only capture domain specific, but struggle to capture domain agnostic patterns. \textbf{(3) DP-DGAD's performance increases with the anomaly ratio.} This trend, also seen in other baselines, occurs because a higher anomaly ratio enables models to capture more diverse anomaly patterns, leading to improved results. Consequently, subsequent experiments will use 10\% anomaly ratio datasets for clearer visualization.

% \begin{table*}[htbp]
% \caption{Ablation study results of DP-DGAD.}
% \label{ablation}
% \renewcommand{\arraystretch}{1.1}
% \centering
% \resizebox{\textwidth}{!}{%
% \begin{tabular}{ll|ccccccccc}
% \hline\hline
% \multirow{2}{*}{\textbf{Metric}} & \multirow{2}{*}{\textbf{Method}} & \multicolumn{8}{c}{\textbf{Dataset}} \\
% \cline{3-10}
% & & AS-Topology & Bitcoin-Alpha & Email-DNC & UCI Messages & Bitcoin-OTC & TAX51 & DBLP & Synthetic-Hijack   \\
% \hline
% \multirow{7}{*}{AUROC}
%     & w/o DPA         & 48.13& 65.54& 60.74& 58.01& 73.27& 43.21& 48.07& 53.68 \\
%     & w/o DPAD       &36.40& 38.72 & 43.95 & 32.40& 49.24& 27.80& 31.59& 44.64\\
%     & w/o CDGA      &50.01&49.24&78.31& 55.17&68.34& 57.42& 47.51&64.74  \\
%     & w/o Wiki &  66.60&46.36&63.39&54.23&48.55&45.86&49.90&65.82\\
%     & w/o Mooc & 58.18&64.88&49.01&63.09&61.08&55.97&33.21&57.43\\
%     & DP-DGAD  &  75.26&81.09&88.87&75.71&90.28&63.21&67.50&91.02\\
% \hline
% \multirow{7}{*}{AUPRC}
%     & w/o DPA         & 20.65& 13.67& 30.99& 20.12& 17.91& 9.32& 9.49& 30.24 \\
%     & w/o DPAD       &7.01& 9.67 & 11.47 & 10.13& 8.44&  5.90& 12.78&13.68\\
%     & w/o CDGA      &21.26&11.10&32.53& 27.86&15.31& 12.67& 7.99&26.71  \\
%     & w/o Wiki &  21.46&12.29&28.22&23.24&15.57&11.44&11.11&27.03\\
%     & w/o Mooc & 19.62&12.08&19.73&29.06&14.97&13.15&9.07&20.77\\
%     & DP-DGAD  &  28.67&18.07&45.17&45.03&24.52&22.23&13.56&35.11\\
% \hline\hline
% \end{tabular}
% }
\begin{table*}[htbp]
\caption{Ablation results of different variants of DP-DGAD.}
\label{ablation}
\centering
\resizebox{\textwidth}{!}{%
\begin{tabular}{ll|cccccccc}
\hline\hline
\multirow{2}{*}{\textbf{Metric}} & \multirow{2}{*}{\textbf{Variants}} & \multicolumn{8}{c}{\textbf{Dataset}} \\
\cline{3-10}
& & AS-Topology & Bitcoin-Alpha & Email-DNC & UCI Messages & Bitcoin-OTC & TAX51 & DBLP & Synthetic-Hijack \\
\hline
\multirow{6}{*}{AUROC}
    & w/o DPA          & 48.13& 65.54& 60.74& 58.01& 73.27& 43.21& 48.07& 53.68 \\
    & w/o DPAD         & 36.40& 38.72 & 43.95 & 32.40& 49.24& 27.80& 31.59& 44.64\\
    & w/o CDGA         & 50.01& 49.24& 78.31& 55.17& 68.34& 57.42& 47.51& 64.74  \\
    & w/o Wiki         & 66.60& 46.36& 63.39& 54.23& 48.55& 45.86& 49.90& 65.82\\
    & w/o MOOC         & 58.18& 64.88& 49.01& 63.09& 61.08& 55.97& 33.21& 57.43\\
    & \textbf{DP-DGAD} & \textbf{75.26}& \textbf{81.09}& \textbf{88.87}& \textbf{75.71}& \textbf{90.28}& \textbf{63.21}& \textbf{67.50}& \textbf{91.02}\\
\hline
\multirow{6}{*}{AUPRC}
    & w/o DPA          & 20.65& 13.67& 30.99& 20.12& 17.91& 9.32& 9.49& 30.24 \\
    & w/o DPAD         & 7.01& 9.67 & 11.47 & 10.13& 8.44&  5.90& 12.78& 13.68\\
    & w/o CDGA         & 21.26& 11.10& 32.53& 27.86& 15.31& 12.67& 7.99& 26.71  \\
    & w/o Wiki         & 21.46& 12.29& 28.22& 23.24& 15.57& 11.44& 11.11& 27.03\\
    & w/o MOOC         & 19.62& 12.08& 19.73& 29.06& 14.97& 13.15& 9.07& 20.77\\
    & \textbf{DP-DGAD} & \textbf{28.67}& \textbf{18.07}& \textbf{45.17}& \textbf{45.03}& \textbf{24.52}& \textbf{22.23}& \textbf{13.56}& \textbf{35.11}\\
\hline\hline
\end{tabular}%
}
\end{table*}
\begin{table*}[htbp]
\caption{Different confidence generation strategies' performance.}
\label{str}
\centering
\resizebox{\textwidth}{!}{%
\begin{tabular}{ll|cccccccc}
\hline\hline
\multirow{2}{*}{\textbf{Metric}} & \multirow{2}{*}{\textbf{Strategies}} & \multicolumn{8}{c}{\textbf{Dataset}} \\
\cline{3-10}
& & AS-Topology & Bitcoin-Alpha & Email-DNC & UCI Messages & Bitcoin-OTC & TAX51 & DBLP & Synthetic-Hijack \\
\hline
\multirow{6}{*}{AUROC}
    & Random Selection          & 52.73 & 58.21 & 64.13& 53.48 & 48.06& 45.18 & 48.92 & 65.33 \\
    & Threshold Based         & 61.42 & 65.89 & 71.08 & 60.95 & 65.78& 52.76 & 55.84 & 73.81\\
    & Distance Based         & 41.95 & 49.67 & 51.01& 45.23 & 50.04& 38.92 & 41.36 & 52.45  \\
    & Similarity Based         & 38.17 & 43.52 & 42.60& 41.66 & 52.12& 35.21 & 37.58 & 49.72\\
    & \textbf{DP-DGAD} & \textbf{75.26}& \textbf{81.09}& \textbf{88.87}& \textbf{75.71}& \textbf{90.28}& \textbf{63.21}& \textbf{67.50}& \textbf{91.02}\\
\hline
\multirow{6}{*}{AUPRC}
    & Random Selection          & 18.94 & 11.85 & 20.38& 25.67 & 9.97& 13.76 & 8.42 & 22.54 \\
    & Threshold Based         & 22.13 & 13.24 & 14.48 & 31.85 & 8.77& 16.89 & 10.38 & 27.93\\
    & Distance Based         & 15.67 & 8.96 & 8.54& 19.22 & 9.09& 10.45 & 6.73 & 18.76  \\
    & Similarity Based         & 13.28 & 7.43 & 5.43& 16.94 & 8.10& 8.92 & 5.21 & 15.48\\
    & \textbf{DP-DGAD} & \textbf{28.67}& \textbf{18.07}& \textbf{45.17}& \textbf{45.03}& \textbf{24.52}& \textbf{22.23}& \textbf{13.56}& \textbf{35.11}\\
\hline\hline
\end{tabular}%
}
\end{table*}
 % Add vertical space between tables

% \resizebox{\textwidth}{!}{%
% \begin{tabular}{ll|ccccccccc}
% \hline\hline
% \multirow{2}{*}{\textbf{Metric}} & \multirow{2}{*}{\textbf{Method}} & \multicolumn{8}{c}{\textbf{Dataset}} \\
% \cline{3-10}
% & & AS-Topology & Bitcoin-Alpha & Email-DNC & UCI Messages & Bitcoin-OTC & TAX51 & DBLP & Synthetic-Hijack   \\
% \hline
% \multirow{7}{*}{AUPRC}
%     & w/o DPA         & 20.65& 13.67& 30.99& 20.12& 17.91& 9.32& 9.49& 30.24 \\
%     & w/o DPAD       &7.01& 9.67 & 11.47 & 10.13& 8.44&  5.90& 12.78&13.68\\
%     & w/o CDGA      &21.26&11.10&32.53& 27.86&15.31& 12.67& 7.99&26.71  \\
%     & w/o Wiki &  21.46&12.29&28.22&23.24&15.57&11.44&11.11&27.03\\
%     & w/o Mooc & 19.62&12.08&19.73&29.06&14.97&13.15&9.07&20.77\\
%     & DP-DGAD  &  28.67&18.07&45.17&45.03&24.52&22.23&13.56&35.11\\
% \hline\hline
% \end{tabular}
% }
% \end{table*}

\subsection{Ablation Study}

\textbf{Effectiveness of Important Module.} In this section, we conduct experiments by removing key components in DP-DGAD to study their effectiveness: (1) \textbf{w/o DPAS} replaces the dynamic prototype-based anomaly scoring with a normal binary classifier. (2) \textbf{w/o DPAD} sets $\lambda_e$ to 0 and $\lambda_d$ to 1, thereby removing the domain adaptation module from the domain adaptive anomaly scoring in DP-DGAD. (3) \textbf{w/o CDGA} removes the confident detection guided adaptation on the target dataset, using the model pretrained on source datasets and tested directly on target datasets. (4) \textbf{w/o Wiki} pretrains DP-DGAD on MOOC only. (5) \textbf{w/o MOOC} pretrains DP-DGAD on Wiki only. 

From the visualization of ablation study in Table~\ref{ablation}, we have the following observations. \textbf{(1) DP-DGAD captures domain-specific pattern, resulting in better performance.} Compared to the ablation variants w/o DPAS, w/o DPAD, and w/o CDGA, the full model demonstrates superior generalizability. This is because dynamic prototype-based anomaly scoring continuously evolves as the prototype buffer updates. As a result, it adapts better to new domains by capturing changing patterns over time. In contrast, w/o DPAS uses a standard binary classifier. This classifier only captures domain-agnostic patterns and cannot evolve to learn domain-specific patterns. As a result, it performs poorly on most datasets compared to DP-DGAD. Similarly, w/o DPAD cannot adapt efficiently to new domains since it lacks the $\lambda_e$ parameter that helps incorporate new domain patterns. Meanwhile, w/o CDGA fails to capture target domain patterns as it must rely solely on domain-agnostic patterns. Consequently, both variants perform poorly compared to DP-DGAD. \textbf{(2) Multiple source datasets help capture domain agnostic patterns for better generalizability.} DP-DGAD is pretrained on both Mooc and Wiki datasets, unlike w/o Wiki and w/o MOOC which use single source datasets. By learning from multiple sources, DP-DGAD can retain shared prototypes that capture common anomalous patterns. These shared prototypes exhibit high similarity to anomalies across different domains, enabling effective generalization to new datasets. Consequently, DP-DGAD generalizes better to target datasets and achieves superior performance compared to single-source variants. More ablation studies can be found in Appendix~\ref{C1}.

\subsection{Pseudo Labeling Strategy Analysis}
\textbf{Comparison of Different Confident Detection Generation Methods.} Apart from important module, we also compare alternative strategies for selecting confident detection for pseudo-labels: \textbf{(1) Random Selection} randomly selects $N_{con}$ pairs from the detection results. \textbf{(2) Threshold Based} uses fixed probability thresholds, selecting pairs with normal detection > $0.7$ and normal detection < $0.3$.  \textbf{(3) Distance Based} selects pairs with maximum Euclidean distance. \textbf{(4) Similarity Based} selects the pairs having lowest Euclidean distance to target datasets.  

From the visualization of different strategies performance in Table~\ref{str}, we have the following observation. \textbf{(1) DP-DGAD generate more reliable psuedo labels, resulting in better performance.} DP-DGAD achieves 15-40\% higher AUROC by selecting low-entropy (high-confidence) detections for pseudo-labeling. In contrast, distance-based and similarity-based methods select pairs based on geometric properties. However, they neglect detection reliability, introducing noise that degrades performance. These geometric methods perform 35-55\% worse than DP-DGAD, often underperforming even random selection. This demonstrates that spatial relationships in latent space are unreliable indicators for pseudo-label quality. Threshold-based selection, using fixed probability cutoffs, achieves 75-85\% of DP-DGAD's effectiveness. While less sophisticated, it captures detection confidence. However, fixed thresholds cannot adapt to varying confidence distributions, unlike DP-DGAD's dynamic entropy-based approach.

\begin{figure}[h!]
\centering
\includegraphics[width=12.7cm]{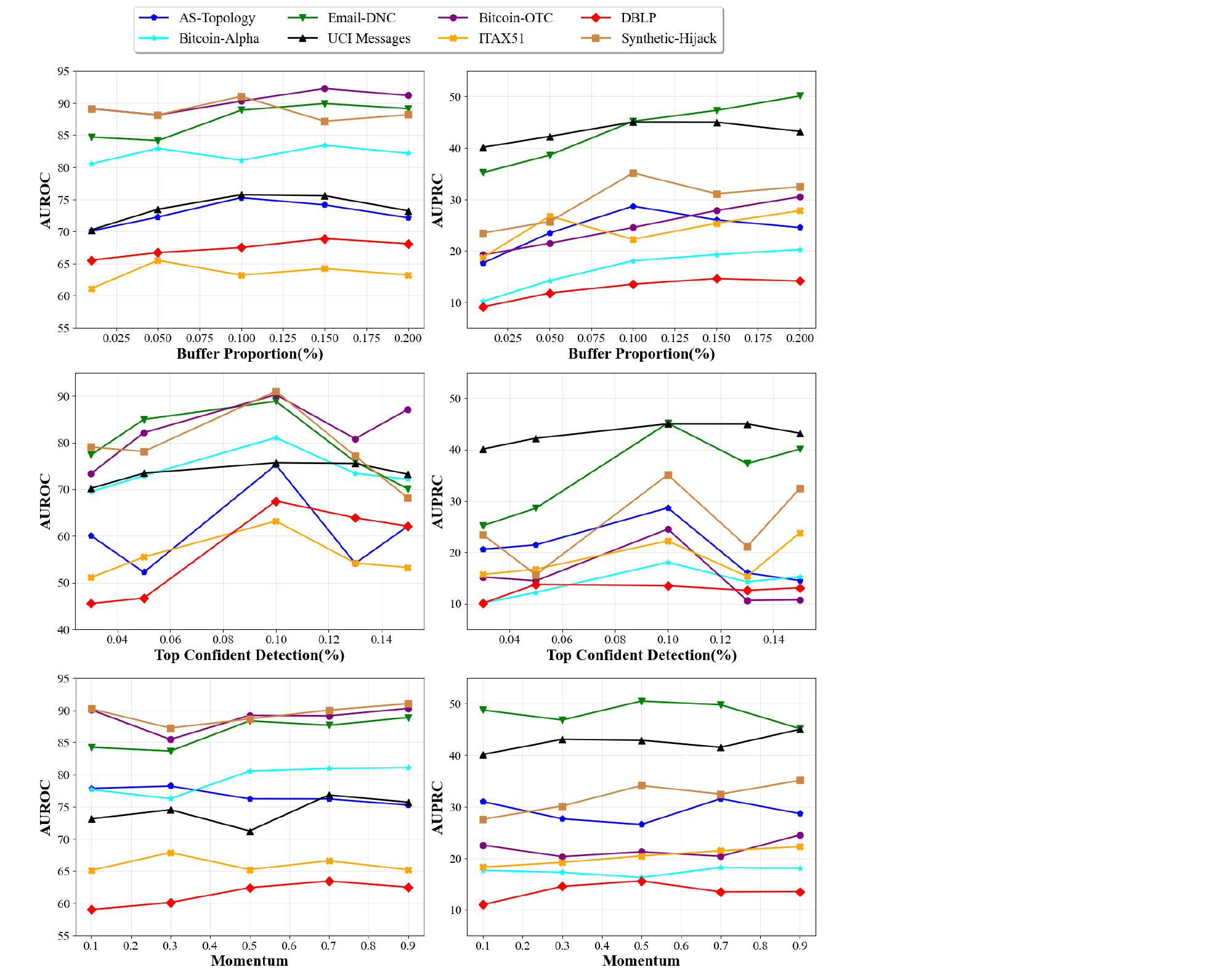}
\caption{\textbf{Impact of dynamic prototype memory buffer size $\mathcal{M}$, confident detection $N_{con}$ and momentum $\alpha$ on the DP-DGAD performance}. } \label{dpm}
\vspace{-0.2cm}
\end{figure}

\subsection{Hyperparameter Analysis}
To analyze DP-DGAD's sensitivity to hyperparameters, we examine three key parameters: the dynamic prototype memory buffer size $\mathcal{M}$, the momentum $\alpha$ and the top confident detection number $N_{con}$ used for pseudo labeling (Here we set it as different proportion to the full dataset size). We vary these values and observe their impact on model performance in Fig.~\ref{dpm}. Our findings are as follows. \textbf{(1) Increasing memory buffer size will improve performance.} 
This is due to the reason that a larger memory buffer will include more prototype pairs, thus covering broader aspects of the distribution of animal patterns. However, we observe an intriguing pattern on some datasets, that Larger buffer sizes actually degrade performance. This occurs because oversized buffers may include irrelevant prototype pairs, which are neither similar nor distinct enough from new domains. They introduce bias and interfere with anomaly scoring, which relies on buffer prototypes to extract distributions. \textbf{(2) Using the top 10\% most confident detections appears to be the optimal setting.}
On most datasets, performance improves as the top confident detection number increases. However, performance begins to drop when this percentage exceeds 10\%. This decline occurs because including more than the top 10\% of detections reduces reliability. These less confident detections have high entropy and introduce additional noise into the pseudo labeling process. \textbf{(3) DP-DGAD remains stable under different momentum.} We vary the momentum value and find DP-DGAD remains stable across all datasets with no clear trend as momentum increases. This stability arises from the fact that memory buffer already captures sufficient similar abnormal and normal patterns through the selection score $s_r$ constraint. As a result, the extracted distributions remain stable regardless of $\alpha$'s value.

\begin{figure}[htbp]
\centering
\includegraphics[width=8.5cm]{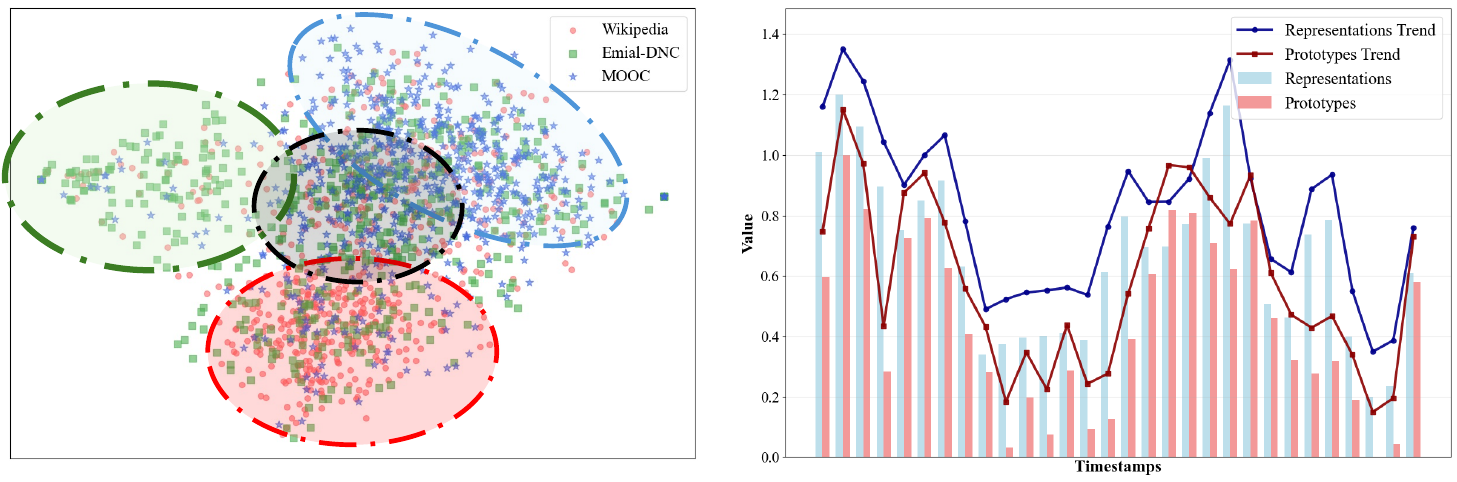}
\caption{\textbf{Visualization of prototypes in memory buffer and how it evolve over time}. } \label{vis}
\end{figure}
\subsection{Visualization}
We visualize dynamic  prototypes from DP-DGAD collected during the pretrain on MOOC, Wikipedia and update Email-DNC. Besides, we also show how dynamic prototypes and edge representations evolve during the training (Fig.~\ref{vis}). 

Our key observations are: (1) Dynamic prototypes capture both domain-specific patterns (three distinct regions dominated by each dataset's prototypes) and domain-agnostic patterns (central region with mixed prototypes from all datasets). (2) Prototypes evolve similarly to representations, demonstrating their temporal adaptability in capturing anomaly patterns.
\section{Conclusion}

In this paper, we propose a generalist DGAD detector that is trained on labeled source datasets and then generalized to target datasets from new domains without labels. Specifically, we aim to capture the evolving anomalous patterns that are both domain agnostic and domain specific, thereby achieving better generalizability compared to existing methods. To this end, we propose DP-DGAD, a generalist detector that leverages dynamic prototypes from temporal ego-graphs to capture evolving patterns. Additionally for unlabeled target datasets, we employ confident pseudo labels to align prototypes with anomalous patterns, enabling effective generalization. Extensive experiments on 10 real-world datasets from various domains demonstrate the effectiveness of DP-DGAD in addressing the generalist DGAD problem. 

\bibliographystyle{ACM-Reference-Format}
\bibliography{sample-base}

\appendix
% \section*{Appendix Contents}

\section{Detailed Experiments Description}

\subsection{Datasets}\label{A1}
For dataset with no groundtruth anomalies. We follow previous work~\cite{liu2021anomaly} to inject anomalies in random timestamps of the dataset. Specifically, take $p$ to be the proportion of total $m$ number of samples in the dataset. We link $p \times m$ number of original disconnected nodes to be the injected anomalies. Details of datasets are as follows:

\noindent\textbf{Wikipedia \cite{kumar2019predicting}}: The Wikipedia dataset consists of one month of edits made on Wikipedia pages. It selects 1,000 most edited pages and editors who made at least 5 edits (a total of 8,227 users), generating 157,474 interactions. Anomalies within this dataset are identified based on public ground-truth labels of banned users, resulting in 217 positive (anomalous) labels.

\noindent\textbf{Bitcoin-OTC and Bitcoin-Alpha \cite{kumar2016edge}}: The Bitcoin-Alpha and Bitcoin-OTC datasets, collected from two Bitcoin platforms (Alpha and OTC), represent users as nodes and trust ratings between them as edges. These ratings, originally on a scale from -10 (fraudster) to +10 (high trust), are scaled to -1 to 1.

\noindent\textbf{Email-DNC \cite{rossi2015network}}: The Email-DNC dataset is a network of emails from the 2016 Democratic National Committee leak, where nodes represent individuals and edges denote email relationships between each individuals.

\noindent\textbf{UCI Messages \cite{zhang2005collecting}}: The UCI Messages dataset, collected from an online community platform, represents users as nodes and messages exchanged between them as edges.

\noindent\textbf{AS-Topology \cite{opsahl2009clustering}}: AS-Topology is a network connection dataset collected from autonomous systems of the Internet. Nodes represent autonomous systems, and edges denote connections between these autonomous systems.

\noindent\textbf{MOOC \cite{kumar2019predicting}}: The MOOC dataset captures student online course related actions, such as viewing videos or submitting answers. It comprises 7,047 users interacting with 98 course items, totaling over 411,749 interactions, and includes 4,066 recorded drop-out events recorded as anomalies.

\noindent\textbf{Synthetic-Hijack \cite{lee2024slade}}: The Synthetic-Hijack dataset is a modified version of the Email-EU dataset, which originally captures email exchanges within a large European research institution. In Synthetic-Hijack dataset, certain normal user accounts are "hijacked" at a specific point in time. These accounts then begin to exhibit anomalous behavior, such as sending spam emails to various recipients. 

\noindent\textbf{TAX51 \cite{xu2023cldg}}: The TAX51 dataset, derived from tax transaction networks, represents companies as nodes and their transaction relationships as edges. 

\noindent\textbf{DBLP \cite{yang2012defining}}: The DBLP collaboration network is a dataset where authors are represented as nodes, and an edge exists between authors who have co-authored a paper. Publication venues (conferences) serve as ground-truth communities, acting as proxies for highly overlapping scientific communities within the network.

\subsection{Baselines}\label{A2}
We give details of all used detection baselines here:

\noindent\textbf{GeneralDyG \cite{yang2025generalizable}}: For fair comparison, we modify GeneralDyG that originally under unsupervised learning setting by adding additional transformer module to compare with abnormal neighbors in labeled source datasets. Then it was updated using only normal samples on target datasets before testing.

\noindent\textbf{FALCON \cite{chen2024fine}}: Originally unsupervised, assuming only normal training data, FALCON was modified by integrating label information into its contrastive learning module alongside temporal patterns for better distinguishability in source datasets. Finally, it was updated using only normal samples on target datasets before testing.

\noindent\textbf{SLADE \cite{lee2024slade}}: On source datasets, SLADE integrates a classification BCE loss with its self-supervised learning to enhance the discriminability. Then, it adapts its learned normal patterns solely through self-supervision using normal samples on target datasets.

\noindent\textbf{SAD \cite{tian2023sad}}: On labeled source datasets, SAD integrates its deviation loss with label enabled contrastive learning to enhance discriminability. While on target datasets, it adapts solely by updating the memory bank and applying pseudo-label contrastive learning to refine normal patterns.

\noindent\textbf{TADDY \cite{liu2021anomaly}}: Here we follow the essence of TADDY to represent dynamic graph using GNN and Transformer for embedding, then guided it with BCE loss on source datasets. It is updated on target dataset using a similar confident detection based psuedo labeling strategy as DP-DGAD.

\noindent\textbf{ARC \cite{liu2024arc}}: To adapt ARC for dynamic graphs, we add additional transformer temporal aware encoder. For labeled source datasets, cross-attention module would leverage both normal and abnormal labels to learn robust anomaly scoring. On unlabeled target datasets, assuming all samples are normal, ARC would continuously refine its internal representation of normal dynamic patterns via the cross-attention module.

\noindent\textbf{AnomalyGFM \cite{qiao2025anomalygfm}}: To adapt AnomalyGFM for dynamic graphs, we add additional transformer temporal aware encoder. For pre-training on labeled source dynamic datasets, AnomalyGFM would learn graph-agnostic normal and abnormal prototypes from these temporal residuals, utilizing both its alignment and classification losses. For unlabeled target dynamic datasets, assuming all samples are normal, the pre-trained encoder and abnormal prototype would be frozen, while the normal prototype would be refined through few-shot prompt tuning using the target dataset's "normal" temporal residuals, enabling robust anomaly scoring.

\noindent\textbf{UNPrompt \cite{niu2024zero}}: For training on labeled source datasets, the backbone model consist of GNN and transformer would be pre-trained via dynamic graph contrastive learning, and neighborhood prompts would learn generalized patterns based on the predictability of a node's latent attributes from its temporal neighbors. The prompts are then finetuned on target datasets.

\noindent\textbf{GraphPrompt \cite{liu2023graphprompt}}: To adapt GraphPrompt for dynamic graph anomaly detection, we replce backbone model with GNN and transformer that can process evolving temporal subgraphs. For labeled source data, a learnable prompt would guide the backbone model to align temporal subgraph representations with normal and abnormal prototypes for anomaly scoring. For unlabeled target data (assumed normal), the prompt would be refined via normal samples to adapt the normal prototype to the new dynamic environment, enabling continuous anomaly detection.

\section{Algorithms}\label{B1}

Algorithms for pretraining on source datasets and updated on target datasets can be found in Algorithm~\ref{alg1} and Algorithm~\ref{alg2}, respectively.

\begin{algorithm}
    \caption{Pretrain \textbf{DP-DGAD} on source datasets}\label{alg1}
    \begin{algorithmic}[1]
    \Statex \textbf{Input:} Source dynamic graph datasets $\{\mathcal{G}_1, \mathcal{G}_2, ..., \mathcal{G}_S\}$; Pre-training epoch $\mathcal{I}$.
    \Statex \textbf{Output:} Function $\Psi_{pretrain}$; $\mathcal{B}$.\\
    Initialize memory buffer $\mathcal{B} \leftarrow \emptyset$.\\
    Initialize GNN and Transformer backbone model.
    \For{datasets $\mathcal{G}_s \in \{\mathcal{G}_1, \mathcal{G}_2, ..., \mathcal{G}_S\}$}
        \For{$i = 1,...,\mathcal{I}$}
            \If {$\mathcal{B}$ is not empty}
                \State Initialize dynamic prototypes $p_n$, $p_a$, $\mu_{n,t}$, $\mu_{a,t}$, $\Sigma_{n,t}$, $\Sigma_{a,t}$ from buffer $\mathcal{B}$
            \Else
                \State Initialize dynamic prototypes $p_n$, $p_a$, $\mu_{n,t}$, $\mu_{a,t}$, $\Sigma_{n,t}$, $\Sigma_{a,t}$ randomly
            \EndIf
            \For{$t \in \{1, 2, ..., T\}$}
                \For{$e_i \in E_t$}
                    \State Extract $k$-hop temporal ego-graph $\mathcal{G}_{ego}$ around $e_i$ 
                    \State Obtain $h_i$ using Eq.1, normalize it using Eq.2
                    \State Get representation $z_i$ of $h_i$ via Eq.3
                    \State Compute anomaly scores $s_i = s_{a,i} - s_{n,i}$ using Eq.17 to Eq.18, then get $L_{BCE}$
                \EndFor
            \EndFor
            \State Align $p_n$, $p_a$ with $\mathbf{Z}_n$, $\mathbf{Z}_a$ using $L_A$ in Eq.7
            \State Update prototype statistics $\mu_{n,t}$, $\mu_{a,t}$, $\Sigma_{n,t}$, $\Sigma_{a,t}$ using Eq.11 to Eq.16
            
            \State Update model using BCE loss $L_{BCE}$ and Alignment loss $L_{A}$
            \If{$\mathcal{G}_s$ is not first source dataset}
                \State Compute similarity score $s_e$ using Eq.9
                \State Compute difference score $s_d$ using Eq.7
                \State Compute ranking score $s_r = \lambda_d \cdot s_d - \lambda_e \cdot s_e$
                \If{$\mathcal{B}$ is full}
                    \If{$s_r >$ lowest $s_r$ in $\mathcal{B}$}
                        \State replace the lowest pair with new $p_n$, $p_a$
                    \Else
                        \State Maintain $\mathcal{B}$
                    \EndIf
                \Else
                    \State Add $p_n$, $p_a$ to $\mathcal{B}$
                \EndIf
            \Else
                \State Compute difference score $s_d$ using Eq.7
                \If{$\mathcal{B}$ is full}
                    \If{$s_d >$ lowest $s_d$ in $\mathcal{B}$}
                        \State replace the lowest pair with new $p_n$, $p_a$
                    \Else
                        \State Maintain $\mathcal{B}$
                    \EndIf
                \Else
                    \State Add $p_n$, $p_a$ to $\mathcal{B}$
                \EndIf
            \EndIf
        \EndFor
    \EndFor
    \end{algorithmic}
\end{algorithm}
\begin{algorithm}
    \caption{Update \textbf{DP-DGAD} on target datasets}\label{alg2}
    \begin{algorithmic}[1]
    \Statex \textbf{Input:} Target dynamic graph datasets $\{\mathcal{G}_1, \mathcal{G}_2, ..., \mathcal{G}_K\}$; Training epoch $\mathcal{I}$; Function $\Psi_{pretrain}$; $\mathcal{B}$.
    \Statex \textbf{Output:} Updated Function $\Psi_{target}$.\\
    \For{$t \in \{1, 2, ..., T\}$}
        \For{$e_i \in E_t$}
            \State Extract $k$-hop temporal ego-graph and get representation $z_i$
            \State Obtain $\mu_{n,t}$, $\mu_{a,t}$, $\Sigma_{n,t}$, $\Sigma_{a,t}$ from buffer $\mathcal{B}$
            \State Compute anomaly scores $S_i = S_{a,i} - S_{n,i}$ using Eq.17 to Eq.18
            \State Convert $s_i$ to $p_i$ and compute $\mathcal{H}(p_i)$
        \EndFor
        \State Select top $N_{con}$ lowest entropy edges, whose detection result will serve as pseudo label
    \EndFor
    \State Initialize dynamic prototypes $p_n$, $p_a$ from buffer $\mathcal{B}$
    \For{$i = 1,...,\mathcal{I}$}
        \For{each confident edge with pseudo-label}
            \State Get representation $z_i$
        \EndFor
        \State Update prototypes using $L_A$, update $\mu_{n,t}$, $\mu_{a,t}$, $\Sigma_{n,t}$, $\Sigma_{a,t}$
        \State Update $\mathcal{B}$ with $s_r$ and $p_n$, $p_a$
    \EndFor

    \end{algorithmic}
\end{algorithm}

\section{Additional Experiments Results}\label{C1}

\begin{table*}[ht!]
\caption{Different proportion of loss parameter's impact on performance.}
\label{loss}
\centering
\resizebox{\textwidth}{!}{%
\begin{tabular}{ll|cccccccc}
\hline\hline
\multirow{2}{*}{\textbf{Metric}} & \multirow{2}{*}{\textbf{Proportion}} & \multicolumn{8}{c}{\textbf{Dataset}} \\
\cline{3-10}
& & AS-Topology & Bitcoin-Alpha & Email-DNC & UCI Messages & Bitcoin-OTC & TAX51 & DBLP & Synthetic-Hijack \\
\hline
\multirow{6}{*}{AUROC}
    & $\lambda_A = 1, \lambda_{BCE} = 0$         & 42.67 & 51.23 & 48.91 & 45.78 & 52.34 & 41.56 & 39.67 & 58.12 \\
    & $\lambda_A = 0.9, \lambda_{BCE} = 0.1$         & 57.82 & 64.56 & 62.13 & 69.45 & 64.91 & 55.34 & 51.78 & 73.29 \\
    & $\lambda_A = 0.7, \lambda_{BCE} = 0.3$         & 63.45 & 70.12 & 73.67 & 60.89 & 66.23 & 59.67 & 53.12 & 77.45 \\
    & $\lambda_A = 0.5, \lambda_{BCE} = 0.5$         & 67.13 & 66.89 & 64.91 & 63.23 & 76.78 & 60.45 & 65.89 & 78.12 \\
    & $\lambda_A = 0.3, \lambda_{BCE} = 0.7$         & 67.67 & 75.34 & 67.56 & 64.12 & 79.45 & 62.89 & 56.23 & 80.78 \\
    & \textbf{$\lambda_A = 0.1, \lambda_{BCE} = 0.9$ (DP-DGAD)} & \textbf{75.26}& \textbf{81.09}& \textbf{88.87}& \textbf{75.71}& \textbf{90.28}& \textbf{63.21}& \textbf{67.50}& \textbf{91.02}\\
    & $\lambda_A = 0, \lambda_{BCE} = 1$         & 31.34 & 48.67 & 40.23 & 52.89 & 50.45 & 38.91 & 47.23 & 57.56 \\
\hline
\multirow{6}{*}{AUPRC}
    & $\lambda_A = 1, \lambda_{BCE} = 0$         & 7.23 & 8.56 & 15.67 & 12.34 & 9.78 & 6.89 & 5.67 & 11.23 \\
    & $\lambda_A = 0.9, \lambda_{BCE} = 0.1$          & 13.89 & 12.34 & 29.23 & 20.12 & 18.67 & 10.45 & 9.78 & 19.67 \\
    & $\lambda_A = 0.7, \lambda_{BCE} = 0.3$          & 14.56 & 16.67 & 21.89 & 21.34 & 11.34 & 18.12 & 11.45 & 21.23 \\
    & $\lambda_A = 0.5, \lambda_{BCE} = 0.5$         & 10.78 & 16.23 & 32.34 & 23.67 & 22.89 & 12.78 & 11.89 & 32.45 \\
    & $\lambda_A = 0.3, \lambda_{BCE} = 0.7$         & 17.12 & 13.45 & 34.56 & 24.23 & 13.67 & 15.89 & 13.12 & 30.78 \\
    & \textbf{$\lambda_A = 0.1, \lambda_{BCE} = 0.9$ (DP-DGAD)} & \textbf{28.67}& \textbf{18.07}& \textbf{45.17}& \textbf{45.03}& \textbf{24.52}& \textbf{22.23}& \textbf{13.56}& \textbf{35.11}\\
    & $\lambda_A = 0, \lambda_{BCE} = 1$         & 6.45 & 7.89 & 9.23 & 5.67 & 6.56 & 10.34 & 7.89 & 5.78 \\
\hline\hline
\end{tabular}%
}
\end{table*}

\begin{table*}[ht!]
\caption{Different proportion of $\lambda_e$ and $\lambda_d$'s impact on performance.}
\label{ds}
\centering
\resizebox{\textwidth}{!}{%
\begin{tabular}{ll|cccccccc}
\hline\hline
\multirow{2}{*}{\textbf{Metric}} & \multirow{2}{*}{\textbf{Proportion}} & \multicolumn{8}{c}{\textbf{Dataset}} \\
\cline{3-10}
& & AS-Topology & Bitcoin-Alpha & Email-DNC & UCI Messages & Bitcoin-OTC & TAX51 & DBLP & Synthetic-Hijack \\
\hline
\multirow{6}{*}{AUROC}
    & $\lambda_e = 0.3, \lambda_d = 0.7$         & 48.93 & 62.34 & 70.45 & 57.89 & 62.56 & 48.67 & 53.21 & 75.34 \\
    & $\lambda_e = 0.9, \lambda_d = 0.1$         & 51.45 & 69.87 & 61.23 & 51.34 & 64.12 & 40.45 & 51.89 & 73.67 \\
    & $\lambda_e = 0.5, \lambda_d = 0.5$         & 50.12 & 70.45 & 63.67 & 53.56 & 67.89 & 47.34 & 55.45 & 68.91 \\
    & $\lambda_e = 0.1, \lambda_d = 0.9$         & 42.67 & 68.23 & 75.12 & 59.23 & 66.34 & 51.89 & 60.78 & 67.23 \\
    & \textbf{$\lambda_e = 0.7, \lambda_d = 0.3$ (DP-DGAD)} & \textbf{75.26}& \textbf{81.09}& \textbf{88.87}& \textbf{75.71}& \textbf{90.28}& \textbf{63.21}& \textbf{67.50}& \textbf{91.02}\\
\hline
\multirow{6}{*}{AUPRC}
    & $\lambda_e = 0.3, \lambda_d = 0.7$          & 8.56 & 6.23 & 18.91 & 21.78 & 10.34 & 9.67 & 6.23 & 11.45 \\
    & $\lambda_e = 0.9, \lambda_d = 0.1$          & 9.12 & 7.89 & 21.34 & 19.45 & 8.78 & 11.12 & 5.67 & 19.78 \\
    & $\lambda_e = 0.5, \lambda_d = 0.5$         & 11.89 & 5.45 & 22.67 & 13.12 & 7.45 & 8.34 & 7.89 & 13.89 \\
    & $\lambda_e = 0.1, \lambda_d = 0.9$         & 12.34 & 9.78 & 20.23 & 12.67 & 9.89 & 10.78 &7.45 & 12.67 \\
    & \textbf{$\lambda_e = 0.7, \lambda_d = 0.3$ (DP-DGAD)} & \textbf{28.67}& \textbf{18.07}& \textbf{45.17}& \textbf{45.03}& \textbf{24.52}& \textbf{22.23}& \textbf{13.56}& \textbf{35.11}\\
\hline\hline
\end{tabular}%
}
\end{table*}

To further analyze DP-DGAD's sensitivity to hyperparameters, we examine two parameter pairs: the loss parameter $\lambda_A$ and $\lambda_{BCE}$, the similarity and difference score parameter $\lambda_d$ and $\lambda_e$. We vary these parameter and visualize them in Table \ref{loss} and Table \ref{ds}. From these two tables we have following observations: \textbf{(1) Balanced alignment and classification losses are crucial for effective prototype learning.} The optimal configuration ($\lambda_A = 0.1$, $\lambda_{BCE} = 0.9$) significantly outperforms both extreme cases. Pure alignment ($\lambda_A = 1$) yields unacceptable AUROC (39.67-58.12\%), lacking BCE's discriminability guidance. Conversely, pure BCE ($\lambda_A = 0$) underperforms (47.23-67.56\% AUROC) as prototypes fail to capture meaningful anomaly patterns without alignment. This highlights that excessive alignment hinders learning, while its complete absence prevents effective prototype-based scoring. The optimal balance ensures prototypes capture anomaly patterns and the embedding space is properly optimized. \textbf{(2) Cross-domain similarity outweighs within-domain distinctiveness for generalization.} DP-DGAD achieves optimal performance with $\lambda_e = 0.7$ and $\lambda_d = 0.3$, prioritizing similarity to new domains over prototype distinctiveness. This configuration enables the model to retain prototypes that generalize well by identifying shared anomalous patterns across domains. The similarity score ($s_e$) is crucial for domain adaptation. Conversely, overemphasizing the difference score ($s_d$) with higher $\lambda_d$ values leads to selecting highly distinct but domain-specific prototype pairs that fail to transfer effectively, causing performance drops of 15-40\% across datasets. This balance ensures DP-DGAD captures domain-agnostic patterns essential for cross-domain generalization while maintaining sufficient discriminative power.
\end{document}